\documentclass{article}
\usepackage{ijcai18}

\usepackage{amsmath}
\usepackage{times}
\usepackage{xcolor}
\usepackage{soul}
\usepackage[utf8]{inputenc}
\usepackage[small]{caption}
\usepackage{CJK}

 \usepackage{mathrsfs}
 \usepackage{algorithm}

 \usepackage{graphicx}
 \usepackage{subfigure}
\usepackage{algorithmic}
\usepackage{multirow}
\usepackage{xcolor}
\usepackage{setspace}
\usepackage{amsfonts}
\usepackage{colortbl,booktabs}
\usepackage{float}
\usepackage{amsfonts,amssymb}

\title{Doubly Aligned Incomplete Multi-view Clustering}

\author{
$\textbf{Menglei Hu}$
\and
$\textbf{Songcan Chen}^{*}$
\\
College of Computer Science \& Technology, Nanjing University of Aeronautics \& Astronautics \\
Collaborative Innovation Center of Novel Software Technology and Industrialization \\
\{ml.hu, s.chen\}@nuaa.edu.cn
}

\begin{document}

\maketitle
\begin{abstract}
Nowadays, multi-view clustering has attracted more and more attention. To date, almost all the previous studies assume that views are complete. However, in reality, it is often the case that each view may contain some missing instances. Such incompleteness makes it impossible to directly use traditional multi-view clustering methods. In this paper, we propose a Doubly Aligned Incomplete Multi-view Clustering algorithm (DAIMC) based on weighted semi-nonnegative matrix factorization (semi-NMF). Specifically, on the one hand, DAIMC utilizes the given instance alignment information to learn a common latent feature matrix for all the views. On the other hand, DAIMC establishes a consensus basis matrix with the help of $L_{2,1}$-Norm regularized regression for reducing the influence of missing instances. Consequently, compared with existing methods, besides inheriting the strength of semi-NMF with ability to handle negative entries, DAIMC has two unique advantages: 1) solving the incomplete view problem by introducing a respective weight matrix for each view, making it able to easily adapt to the case with more than two views; 2) reducing the influence of view incompleteness on clustering by enforcing the basis matrices of individual views being aligned with the help of regression. Experiments on four real-world datasets demonstrate its advantages.
\end{abstract}

 \section{Introduction}
Many datasets in real world naturally appear in multiple views or come from multiple sources \cite{blum1998combining,schechter2017processing}, which are called multi-view data. For example, a document can be translated into different languages, and images can be described by different features such as Fourier shape descriptors and K-L expansion coefficients. In multi-view data, these different views often share some consistency and complementary information \cite{sun2013survey,zhao2017multi}. Such information can be beneficial to learning tasks such as classification and clustering. This leads to a surge of interest in multi-view learning \cite{potthast2018active,jing2017semi}, whose goal is to integrate information and give a compatible solution across all views. Nowadays, multi-view learning has been widely studied in different areas such as face recognition, image processing and natural language processing \cite{romero2017multi,xing2017towards,nie2018auto}.

In all the tasks of multi-view learning, multi-view clustering \cite{bickel2004multi,fan2017robust,nie2017multi,chao2017survey} has attracted more and more attentions due to exempting the expensive requirement of data labeling. The goal of multi-view clustering is making full use of multi-view data to get a better clustering result than just simply concatenated views. To date, many multi-view clustering methods have been proposed. Among these methods, one of the most widely used techniques is nonnegative matrix factorization (NMF) \cite{wang2016adaptive,li2016advances}. \cite{lee1999learning} proposes the NMF, which has received much attention because of its straightforward interpretability for applications. Then, some researchers utilize the NMF for multi-view learning, especially multi-view clustering. A joint NMF process with the consistency constraint is formulated in \cite{liu2013multi}, which performs NMF for each view and pushes each view’s low dimensional representation towards a common consensus. Besides, some researchers have integrated manifold learning and multi-view learning by imposing the manifold regularization on the objective function of NMF respectively for individual views data \cite{wang2016multi,zong2017multi}.

Most of the previous studies on multi-view clustering make a common assumption that all of the views are complete. However, in real world applications, multi-view data tend to be incomplete. For example, in the camera network, for some reasons, the camera may temporarily fail, or be blocked by some objects, making the instance missing. Another example is in document clustering, different languages of the documents represent multiple views. However, we may not get all the documents translated into each language. All the above-mentioned cases lead to the incompleteness of multi-view data. As a result, the lack of the whole row or column makes the traditional instance imputation methods fail. So how to make full use of the complementary knowledge hidden in different views and reduce the impact of missing instances are the most challenging problems of incomplete multi-view learning. Recently, some incomplete multi-view clustering methods have been proposed, for example, \cite{li2014partial} proposes PVC by utilizing the information of instance alignment to learn a common latent subspace for aligned instances and a private latent representation for unaligned instances via NMF. Borrowing this idea, \cite{zhao2016incomplete} proposes IMG by integrating PVC and manifold learning to learn the global structure over the entire data instances across all views. However, PVC and IMG can only deal with the two-view incomplete multi-view clustering, limiting their application scope. A method for clustering more than two incomplete views is proposed in \cite{shao2015multiple}(MIC) by  filling the missing instances with the average feature values in each incomplete view, then handling the problem with the help of weighted NMF and $L_{2,1}$-Norm regularization \cite{kong2011robust,wu2018manifold}. However, such a simple imputation will cause a great deviation when the missing ratio is large. As a result, incomplete multi-view clustering still faces significant challenges.

In this paper, we propose the Doubly Aligned Incomplete Multi-view Clustering (DAIMC) to meet the challenges. By integrating semi-NMF \cite{ding2010convex} and \emph{L}$_{2,1}$-Norm regularized regression model, DAIMC tries to learn a common latent feature matrix for all the views from two aspects of instances aligned and the basis matrix aligned.

Compared with the existing methods around the incomplete multi-view clustering,  besides inheriting the strength of semi-NMF with ability to handle negative entries, DAIMC mainly has the following advantages:\\
1. DAIMC extends PVC. Borrowing the idea of the weighted NMF, DAIMC introduces a respective weight matrix for each incomplete view to assign the missing instances zero weights and the presented instances one weights in each view, making it able to be easily and straightforwardly extended to the scenario with more than two incomplete views.\\
2. Besides the instance alignment, DAIMC considers the global information by enforcing the basis matrices of individual views being aligned with the help of the \emph{L}$_{2,1}$-Norm regularized regression model, which further reduces the influence of missing instances on clustering performance.\\
3. An iterative optimization algorithm for DAIMC with convergent guarantee is proposed. Experimental results on four real-world datasets demonstrate its advantages.

The rest of this paper is organized as follows. In Section 2, we overview some related work on semi-NMF and incomplete multi-view clustering. Section 3 proposes our DAIMC and an efficient iterative algorithm of solving it in detail, respectively. Extensive experimental results and analysis are reported in Section 4. Section 5 concludes this paper with future research directions.
\section{Related Work}

\subsection{Semi-nonnegative Matrix Factorization}
The semi-NMF \cite{ding2010convex} is an effective latent factor learning method , which is the extension of NMF. Given an input data matrix $\textbf X\in\mathbb{R}^{M \times N}$, each column of $ \textbf X $ is an instance vector. The semi-NMF aims to find a matrix $ \textbf U\in\mathbb{R}^{M \times K} $ and a nonnegative matrix $ \textbf V\in\mathbb{R}^{N \times K} $ whose product can well approximate the original matrix $ \textbf X $. To facilitate discussion, we call $ \textbf U $ the $basis\ matrix$ and $ \textbf V $ the $latent\ feature\ matrix$. Thus we can get the following minimization problem
\begin{equation}
\begin{aligned}
&\textrm {min}\  \| \textbf X - \textbf U\textbf V^T \|_{\emph{\scriptsize{F}}}^{2}\\
&\textrm {s.t.}\ \ \textbf {V} \geq 0
\end{aligned}
\end{equation}
Similar to NMF, the objective function in (1) is biconvex. Therefore, it is unrealistic to expect an algorithm to find the global minimum. \cite{ding2010convex} proposes an iterative updating algorithm to find the locally optimal solution as follows:\\
Update $ \textbf U $ (while fixing $ \textbf V $) using the rule
\begin{equation}
\begin{aligned}
\textbf U=\textbf X\textbf V(\textbf V^{T}\textbf V)^{-1}
\end{aligned}
\end{equation}
Update $ \textbf V $ (while fixing $ \textbf U $) using
\begin{equation}
\begin{aligned}
\textbf V_{jk} \gets \textbf V_{jk}\sqrt{\frac{(\textbf X^{T}\textbf U)_{jk}^{+} + [\textbf V(\textbf U^{T}\textbf U)^{-}]_{jk}}{(\textbf X^{T}\textbf U)_{jk}^{-} + [\textbf V(\textbf U^{T}\textbf U)^{+}]_{jk}}}
\end{aligned}
\end{equation}
where we separate the positive and negative parts of a matrix $\textbf A$ as:
\begin{equation}
\begin{aligned}
\textbf A_{jk}^{+}=(|\textbf A_{jk}|+\textbf A_{jk})/2,\qquad \textbf A_{jk}^{-}=(|\textbf A_{jk}|-\textbf A_{jk})/2.
\end{aligned}
\end{equation}

\subsection{Incomplete Multi-view Clustering}
Given a dataset with $N$ instances, $C$ categories, $n_{v}$ views $ \{\textbf X^{(i)},i=1,2,...,n_{v}\} $, where $ \textbf X^{(v)}\in\mathbb{R}^{d_{v} \times N} $ is the $v$-th view of the dataset. An indicator matrix $ \textbf M\in\mathbb{R}^{n_{v} \times N} $ for incomplete multi-view clustering problem is defined as:
\begin{eqnarray}
\textbf M_{ij}=
\begin{cases}
1& \text{if $j$-th instance is in the $i$-th view} \\
0& \text{otherwise}
\end{cases}
\end{eqnarray}

where each row of $ \textbf M $ is the instance presence for corresponding view. If every view contains all the instances, then the matrix $ \textbf M $ is an all one matrix. And if the $v$-th view is incomplete, the data matrix $ \textbf X^{(v)} $ will have a number of column missing, $i.e., \sum_{j=1}^{N}\textbf M_{vj}<N.$

The aim of the incomplete multi-view clustering is to integrate all the incomplete views to cluster the $N$ instances into $C$ clusters.

\section{Proposed Approach}
In this section, we present our Doubly Aligned Incomplete Multi-view Clustering(DAIMC) in detail. We model the DAIMC as a joint weighted semi-NMF problem and use $L_{2,1}$-norm regularized regression to enforce the basis matrix of individual views being aligned. In the following, we propose our model in two aspects and then give a unified objective function for implementing DAIMC.

\subsection{Weighted Semi-NMF for Incomplete Multi-view Data}
For the $v$-th view, similarly to the weighted NMF, the weighted semi-NMF factorizes the data matrix $ \textbf X^{(v)}\in\mathbb{R}^{d_{v} \times N} $ into two matrix $\textbf U^{(v)}$ and $\textbf V^{(v)}$, where $ \textbf U^{(v)}\in\mathbb{R}^{d_{v} \times K} $,
$ \textbf V^{(v)}\in\mathbb{R}^{N \times K} $,while giving different weights to the reconstruction errors of different instances. $K$ denotes dimension of subspace. In the experiments of previous works \cite{shao2015multiple,zhao2016incomplete}, for multi-view clustering, the $K$ is set to the number of the categories of the data matrix $ \textbf X^{(v)}$, i.e., $K=C$. As a result, the weighted semi-NMF optimization problem is formulated as:
\begin{equation}
\begin{aligned}
&\textrm {min}\  \| (\textbf X^{(v)} - \textbf U^{(v)} \textbf V^{(v)^T})\textbf W^{(v)} \|_{\emph{\scriptsize{F}}}^{2}\\
&\textrm {s.t.}\ \ \textbf V^{(v)} \geq 0
\end{aligned}
\end{equation}
where the weight matrix $ \textbf W^{(v)}\in\mathbf{R}^{N \times N} $ is a diagonal matrix.
\begin{eqnarray}
\textbf W_{jj}^{(i)}=
\begin{cases}
1& \text{if $j$-th instance is in the $i$-th view} \\
0& \text{otherwise}
\end{cases}
\end{eqnarray}

Note that $\textbf W_{jj}^{(v)}$ indicates the weight of the $j$-th instance in view $v$. If the $j$-th instance is missing, then the loss of this instance will be ignored.

However, (6) only independently decomposes different views without considering their consistency information. To address this issue, we assume that different views have distinct basis matrices $\{\textbf U^{(i)}\}_{i=1}^{n_{v}}$ , but share the same latent feature space $\textbf V$. As a result, (6) is rewritten as follows:
\begin{equation}
\begin{aligned}
&\textrm {min}\  \sum_{i=1}^{n_{v}}\| (\textbf X^{(i)} - \textbf U^{(i)} \textbf V^T)\textbf W^{(i)} \|_{\emph{\scriptsize{F}}}^{2}\\
&\textrm {s.t.}\ \ \textbf V \geq 0
\end{aligned}
\end{equation}
By solving (8), we can obtain a common representation $\textbf V$ for multiple incomplete-view instances.

 \subsection{$L_{2,1}$-Norm Regularized Regression with Basis Matrix}
 To further reduce the influence of the missing instances, DAIMC attempts to incorporate the global information among views. In multi-view data, different views have different representations for a same data matrix. Thus, we can align the different basis matrices $\{\textbf U^{(i)}\}_{i=1}^{n_{v}}$ of individual views with the help of regression by solving the following problem for the basis matrices intended to be aligned.
\begin{equation}
\begin{aligned}
\textrm {min}\  \sum_{i=1}^{n_{v}}\| \textbf B^{(i)^T} \textbf U^{(i)} - \textbf U^{(*)} \|_{\emph{\scriptsize{F}}}^{2} + \beta\| \textbf B^{(i)} \|_{2,1}
\end{aligned}
\end{equation}
where $ \textbf B^{(i)}\in\mathbb{R}^{d_{i} \times K} $ is the regression coefficient matrix for view $i$. The $L_{2,1}$-norm regularization term is here introduced for ensuring $\textbf B^{(i)}$ sparse in rows. In this way, $\textbf B^{(i)}$ performs a feature selection during the alignment process. The matrix $\textbf U^{(*)}\in\mathbb{R}^{P \times K} $ is the same low dimensional representation for the basis matrices of all the views. $P$ denotes dimension of subspace. The value of $P$ will affect the result. Instead of looking for an appropriate $P$, we simply set the matrix $\textbf U^{(*)}$ equal to a $K$ dimensional identity matrix $\textbf I_K$, whose columns correspond to the cluster encodings. For such a setting, our experiments later confirm its effectiveness. As a result, (9) is rewritten as:
\begin{equation}
\begin{aligned}
\textrm {min}\  \sum_{i=1}^{n_{v}}\| \textbf B^{(i)^T} \textbf U^{(i)} - \textbf I \|_{\emph{\scriptsize{F}}}^{2} + \beta\| \textbf B^{(i)} \|_{2,1}
\end{aligned}
\end{equation}
where $\beta$ is the trade-off hyper-parameter between sparsity and accuracy of regression for the $i$-th view, $\|.\|_{2,1}$ is the $L_{2,1}$ norm and defined as:
\begin{equation}
\begin{aligned}
\|\textbf B^{(v)}\|_{2,1}=\sum_{i=1}^{d_{v}}\sqrt{\sum_{j=1}^{K}{\textbf B^{(v)}_{ij}}^{2}}\nonumber
\end{aligned}
\end{equation}
 \subsection{Unified Objective Function}
Considering the objective for instance alignment information as well as the basis matrix alignment information simultaneously, we minimize the following objective function
\begin{equation}
\begin{aligned}
\mathcal{J}=&\sum_{i=1}^{n_{v}}\big\{\| (\textbf X^{(i)} - \textbf U^{(i)} \textbf V^T)\textbf W^{(i)} \|_{\emph{\scriptsize{F}}}^{2}\\
&+ \alpha(\| \textbf B^{(i)^T} \textbf U^{(i)} - \textbf I \|_{\emph{\scriptsize{F}}}^{2} + \beta\| \textbf B^{(i)} \|_{2,1})\big\}\\
\textrm {s.t.}\ &\ \textbf V \geq 0
\end{aligned}
\end{equation}
where $\alpha$ is nonnegative hyper-parameter that controls the trade-off between the aforementioned two objectives.

 \subsection{Optimization}
The objective function in Eq.(11) is not convex over all variables $\{\textbf U^{(i)}\}_{i=1}^{n_{v}}$, $\textbf V$, $\{\textbf B^{(i)}\}_{i=1}^{n_{v}}$, simultaneously. To solve this optimization problem, we propose an alternating iteration procedure.
\\
$\textbf {Subproblem\ of}$ $\{\textbf U^{(i)}\}_{i=1}^{n_{v}}$.
With $\{\textbf B^{(i)}\}_{i=1}^{n_{v}}$ and $\textbf V$ fixed, for each $\textbf U^{(i)}$, we need to minimize the following objective function:
\begin{equation}
\begin{aligned}
\mathcal{J}(\textbf U^{(i)})=\| (\textbf X^{(i)} - \textbf U^{(i)} \textbf V^T)\textbf W^{(i)} \|_{\emph{\scriptsize{F}}}^{2} + \alpha\| \textbf B^{(i)^T} \textbf U^{(i)} - \textbf I \|_{\emph{\scriptsize{F}}}^{2}\\
\end{aligned}
\end{equation}
The partial derivation of $\mathcal{J}(\textbf U^{(i)})$ with respect to $\textbf U^{(i)}$ is
\begin{equation}
\begin{aligned}
\frac{\partial \mathcal{J}}{\partial \textbf U^{(i)}}=&2(\textbf U^{(i)}\textbf V^T-\textbf X^{(i)})\textbf W^{(i)}\textbf W^{(i)^T}\textbf V + \\
&2\alpha\textbf B^{(i)}(\textbf B^{(i)^T}\textbf U^{(i)}-\textbf I)
\end{aligned}
\end{equation}
From the definition of $\textbf W^{(i)}$, we can see $\textbf W^{(i)}=\textbf W^{(i)}\textbf W^{(i)^T}$. Let $\partial \mathcal{J}/\partial \textbf U^{(i)}=0$, we get the following equation:
\begin{equation}
\begin{aligned}
(\textbf U^{(i)}\textbf V^T-\textbf X^{(i)})\textbf W^{(i)}\textbf V + \alpha\textbf B^{(i)}(\textbf B^{(i)^T}\textbf U^{(i)}-\textbf I)=0
\end{aligned}
\end{equation}
Eq.(14) is called the $continuous\ Sylvester\ equation$ with respect to $\textbf U^{(i)}$, which often arises in control theory. When both $d_{i}$ and $K$ are small, we can solve Eq.(14) via vectorization and get the following updating rule for $\textbf U^{(i)}$:
\begin{equation}
\begin{aligned}
vec(\textbf U^{(i)})=&[\textbf I_{K} \otimes (\alpha\textbf B^{(i)}\textbf B^{(i)^T}) + (\textbf V^T\textbf W^{(i)}\textbf V) \otimes \textbf I_{d_{i}}]^{-1}\\
&vec(\textbf X^{(i)} \textbf W^{(i)} \textbf V + \alpha\textbf B^{(i)})
\end{aligned}
\end{equation}
And when both $d_{i}$ and $K$ are large, we instead solve it via $Conjugate\ Gradient$. In our experiments, we use the $lyap$ function of MATLAB to solve Eq.(14).\\
$\textbf {Subproblem\ of}$ $\textbf V$.
With $\{\textbf B^{(i)}\}_{i=1}^{n_{v}}$ and  $\{\textbf U^{(i)}\}_{i=1}^{n_{v}}$ fixed, we need to minimize the following objective function:
\begin{equation}
\begin{aligned}
&\mathcal{J}(\textbf V)=\  \sum_{i=1}^{n_{v}}\| (\textbf X^{(i)} - \textbf U^{(i)} \textbf V^T)\textbf W^{(i)} \|_{\emph{\scriptsize{F}}}^{2}\\
&\textrm {s.t.}\ \ \textbf V \geq 0
\end{aligned}
\end{equation}
The partial derivation of $\mathcal{J}(\textbf V)$ with respect to $\textbf V$ is
\begin{equation}
\begin{aligned}
\frac{\partial \mathcal{J}}{\partial \textbf V}=\  \sum_{i=1}^{n_{v}}\big\{2\textbf W^{(i)}\textbf V\textbf U^{(i)^T}\textbf U^{(i)}-2\textbf W^{(i)}\textbf X^{(i)^T}\textbf U^{(i)}\big\}
\end{aligned}
\end{equation}
Similar to the optimization of semi-NMF, using the KKT complementary condition for the nonnegativity of $\textbf V$, we get
\begin{equation}
\begin{aligned}
(\sum_{i=1}^{n_{v}}\textbf W^{(i)}\textbf V\textbf U^{(i)^T}\textbf U^{(i)}-\sum_{i=1}^{n_{v}}\textbf W^{(i)}\textbf X^{(i)^T}\textbf U^{(i)})_{jk}\textbf V_{jk}=0
\end{aligned}
\end{equation}
Based on this equation, we can write the updating rule for $\textbf V$ as:
\begin{equation}
\begin{aligned}
 &\textbf V_{jk} \gets\textbf V_{jk}\\
 &\sqrt{\frac{\sum_{i=1}^{n_{v}}\big\{[\textbf W^{(i)}(\textbf X^{(i)^T}\textbf U^{(i)})^{+}]_{jk} + [\textbf W^{(i)}\textbf V(\textbf U^{(i)^T}\textbf U^{(i)})^{-}]_{jk}\big\}}{\sum_{i=1}^{n_{v}}\big\{[\textbf W^{(i)}(\textbf X^{(i)^T}\textbf U^{(i)})^{-}]_{jk} + [\textbf W^{(i)}\textbf V(\textbf U^{(i)^T}\textbf U^{(i)})^{+}]_{jk}\big\}}}\\
\end{aligned}
\end{equation}
It is worth to note that if $\{\textbf U^{(i)}\}_{i=1}^{n_{v}}$ and $\textbf V$ are a solution of Eq.(11), then $\{\textbf U^{(i)}\textbf Q\}_{i=1}^{n_{v}}$ and $\textbf V\textbf Q^{-1}$ will form another solution for any invertible matrix $\textbf Q$. With these requirements, the normalization imposed on $\{\textbf U^{(i)}\}_{i=1}^{n_{v}}$ and $\textbf V$ are achieved by
\begin{equation}
\begin{aligned}
&&\textbf V \gets \textbf V\textbf Q^{-1}\\
&&\textbf U^{(i)} \gets \textbf U^{(i)}\textbf Q
\end{aligned}
\end{equation}
where $\textbf Q$ is a diagonal matrix formally defined as\cite{wang2016multi}:
\begin{equation}
\begin{aligned}
\textbf Q = diag(\sum_{i}\textbf V_{i1},\sum_{i}\textbf V_{i2},...,\sum_{i}\textbf V_{iK}).
\end{aligned}
\end{equation}
\\
$\textbf {Subproblem\ of}$ $\{\textbf B^{(i)}\}_{i=1}^{n_{v}}$.
With $\{\textbf U^{(i)}\}_{i=1}^{n_{v}}$ and $\textbf V$ fixed, for each $\textbf B^{(i)}$, we need to minimize the following objective function:
\begin{equation}
\begin{aligned}
\mathcal{J}(\textbf B^{(i)})=\| \textbf B^{(i)^T} \textbf U^{(i)} - \textbf I \|_{\emph{\scriptsize{F}}}^{2} + \beta\| \textbf B^{(i)} \|_{2,1}
\end{aligned}
\end{equation}
The partial derivation of $\mathcal{J}(\textbf B^{(i)})$ with respect to $\textbf B^{(i)}$ is
\begin{equation}
\begin{aligned}
\frac{\partial \mathcal{J}}{\partial \textbf B^{(i)}}= 2\textbf U^{(i)}(\textbf U^{(i)^T}\textbf B^{(i)}-\textbf I) + \beta\textbf D^{(i)}\textbf B^{(i)}
\end{aligned}
\end{equation}
Where $\textbf D^{(i)}$ is a diagonal matrix with the $j$-th diagonal element given by
\begin{equation}
\begin{aligned}
\textbf D^{(i)}_{jj}=\frac{1}{\|\textbf B_{j:}^{(i)}\|_2}
\end{aligned}
\end{equation}
where $\textbf B_{j:}^{(i)}$ is the $j$-th row of matrix $\textbf B^{(i)}$. Let $\partial \mathcal{J}/\partial \textbf B^{(i)}$ = 0, we get the following updating rule for $\textbf B^{(i)}$:
\begin{equation}
\begin{aligned}
\textbf B^{(i)} = [\textbf U^{(i)}\textbf U^{(i)^T}+0.5\beta\textbf D^{(i)}]^{-1}\textbf U^{(i)}
\end{aligned}
\end{equation}
Generally, in real-world dataset, $d_i\gg K$, thus $\textbf U^{(i)}\textbf U^{(i)^T}+0.5\beta\textbf D^{(i)}$ is close to singular. In order to avoid inaccurate results and reduce the complexity of the algorithm, we use the matrix inverse equation \cite{bishop2007pattern}, to reformulate the update rules for $\textbf B^{(i)}$:
\begin{equation}
\begin{aligned}
\textbf B^{(i)} = &\frac{2}{\beta}[\textbf D^{(i)^{-1}} - \textbf D^{(i)^{-1}}\textbf U^{(i)} (\textbf U^{(i)^T}\textbf D^{(i)^{-1}}\textbf U^{(i)} + 0.5\beta\textbf I)^{-1}\\
&\textbf U^{(i)^T}\textbf D^{(i)^{-1}}]\textbf U^{(i)}
\end{aligned}
\end{equation}
The entire optimization procedure for DAIMC is summarized in Algorithm 1.

\begin{algorithm}[!t]
\caption{Optimization of DAIMC}
\label{alg1}
\begin{algorithmic}[1]
\REQUIRE
Data matrices for incomplete views $\textbf X^{(1)},...,\textbf X^{(n_v)}$, weight matrices $\textbf W^{(1)},...,\textbf W^{(n_v)}$,  hyper-parameters $\alpha,\beta$, number of clusters $K$.\\
\STATE Initialize $\textbf V\in\mathbb{R}^{N \times K}$, $\textbf U^{(i)}\in\mathbb{R}^{d_{i} \times K}$ and $\textbf B^{(i)}\in\mathbb{R}^{d_{i} \times K}$ $i=1,...,n_v$.
\REPEAT
\FOR{$i=1$ to $n_v$}
\STATE Update $\textbf U^{(i)}$ by Eq.(14) and $lyap$ function.
\STATE Update $\textbf B^{(i)}$ by Eq.(26).
\ENDFOR
\REPEAT
\STATE Update $\textbf V$ by Eq.(19).
\UNTIL{Eq.(16) converges}
\STATE Normalize $\textbf U^{(i)}$s and $\textbf V$ by Eq.(20).
\UNTIL{Eq.(11) converges}
\STATE Apply $K$-means on $\textbf V$ to get the clustering results.
\RETURN
Basis matrices $\textbf U^{(1)},...,\textbf U^{(n_v)}$, regression coefficient matrices $\textbf B^{(1)},...,\textbf B^{(n_v)}$, the common latent feature matrix $\textbf V$ and clustering results.
\end{algorithmic}
\end{algorithm}
\subsection{Convergence and Complexity}
$Convergence\  Analysis$. As shown by Algorithm 1, the optimization of DAIMC can be divided into three subproblems, each of which is convex w.r.t one variable. Thus, by finding the optimal solution for each subproblem alternatively, our algorithm can at least find a locally optimal solution.
\\
$Complexity\  Analysis$. The time complexity of DAIMC is dominated by matrix multiplication and inverse operations. In each iteration, the $lyap$ function costs $\mathcal{O}(d_i^3)$ and the matrix inversion in Eq.(26) costs $\mathcal{O}(K^3)$. The complexities of multiplication operations in updating $\textbf U^{(i)}$, $\textbf V$ and $\textbf B^{(i)}$ are $\mathcal{O}(d_i^2K+d_iN+KN)$, $\mathcal{O}(M(d_iKN+d_iK^2+K^2N))$ and $\mathcal{O}(K^3+d_i^2K)$ respectively, where $M$ is the iteration times of the inner loop. In general, $K \leq d_{i}$ and $N$. Suppose $L$, $d_{max}$ are the iteration times of the outer loop and the largest dimensionality of all the views respectively, thus the time complexity of DAIMC is $\mathcal{O}(n_vLd_{max}^3+LMd_{max}KN)$.
 \section{Experiments and Analysis}
$\textbf {Datasets}$: The experiments are conducted on four real-world multi-view datasets. The important statistics of these datasets are given in the Table \ref{tab:datasets}.
\begin{table}[htb]
\tabcolsep 5pt
\vspace*{-12pt}
\begin{center}
\def\temptablewidth{0.4\textwidth}
{\rule{\temptablewidth}{1pt}}
\begin{tabular*}{\temptablewidth}{@{\extracolsep{\fill}}ccccccc}
Dataset &$\#$ instances &$\#$ views &$\#$ clusters\\   \hline
Wikipedia\footnotemark[1] & 693 & 2 & 10 \\
Digit\footnotemark[2] & 2000 & 5 & 10 \\
3Sources\footnotemark[3] & 416 & 3 & 6 \\
Flowers & 1360 & 3 & 17  \\
       \end{tabular*}
       {\rule{\temptablewidth}{1pt}}
        \caption{Statistics of the datasets}
        \label{tab:datasets}
       \end{center}
       \end{table}
\footnotetext[1]{http://www.svcl.ucsd.edu/projects/crossmodal/}
\footnotetext[2]{http://archive.ics.uci.edu/ml/datasets.html}
\footnotetext[3]{http://mlg.ucd.ie/datasets/3sources.html}
\\
\begin{figure*}[htb]
      \centering
      \subfigure[ACs for Wikipedia]{
           \label{plot:ACs for Wikipedia}
           \includegraphics[width=0.29\textwidth]{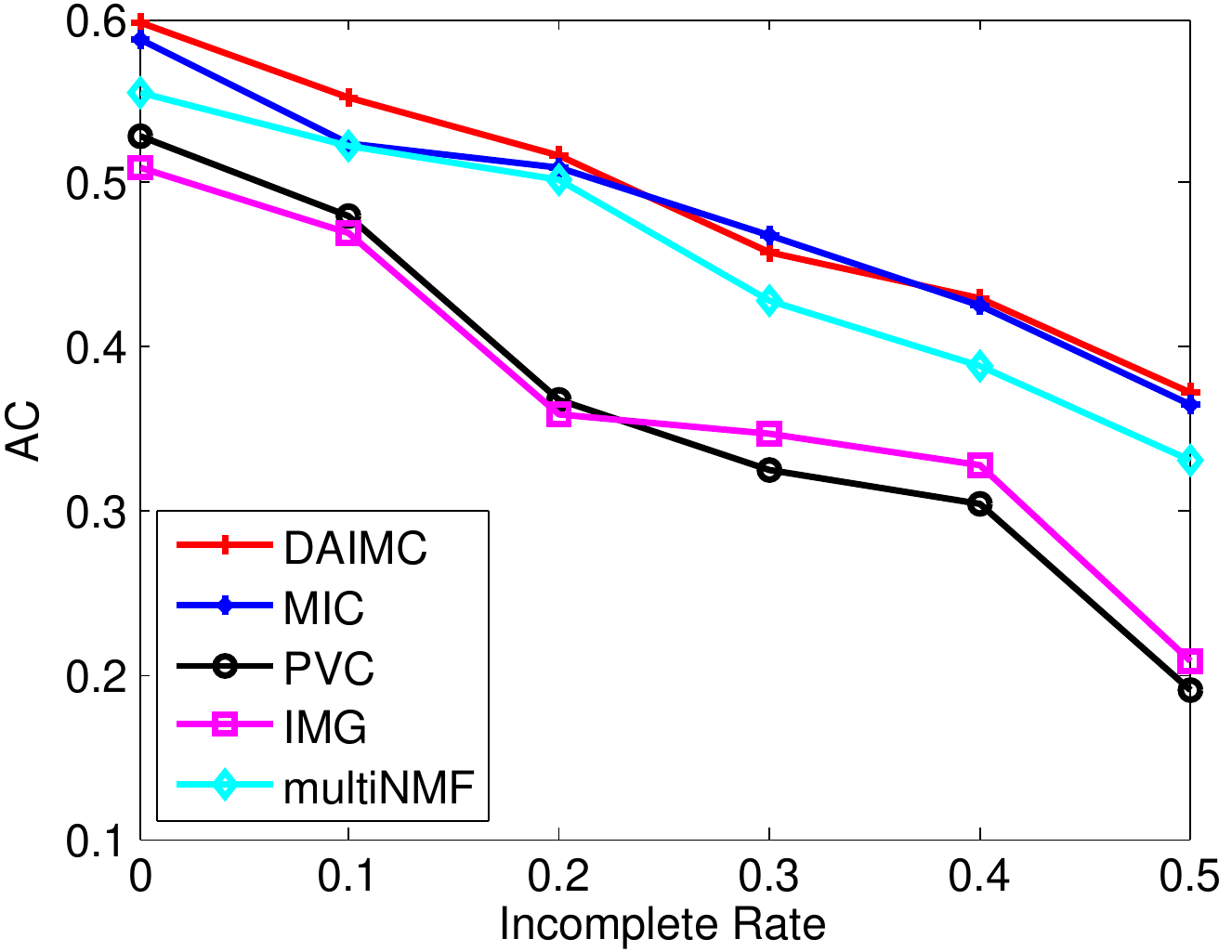}
      }
      \subfigure[ACs for Digit]{
           \label{plot:ACs for Digit}
           \includegraphics[width=0.29\textwidth]{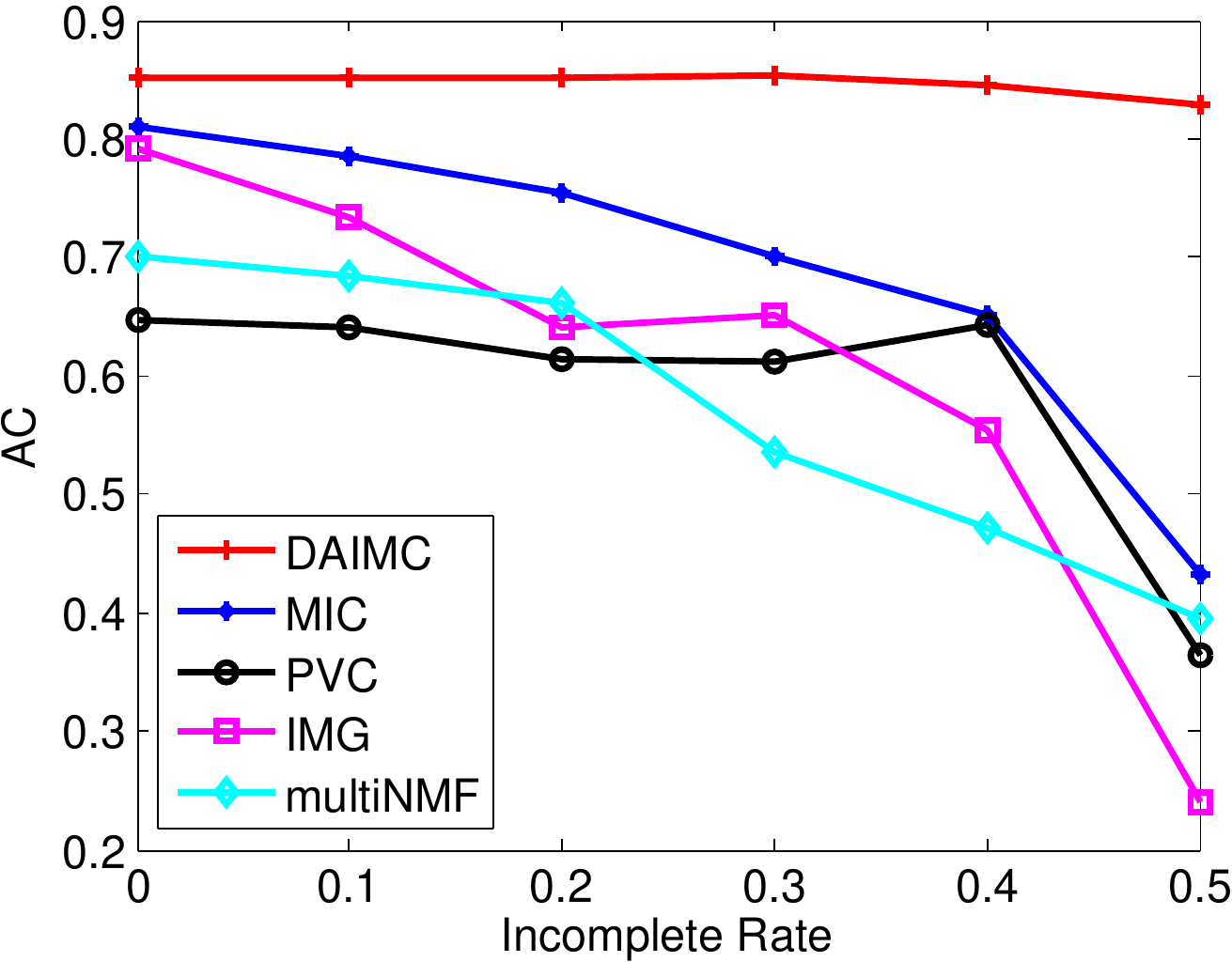}
      }
      \subfigure[ACs for Flowers]{
           \label{plot:ACs for Flowers}
           \includegraphics[width=0.29\textwidth]{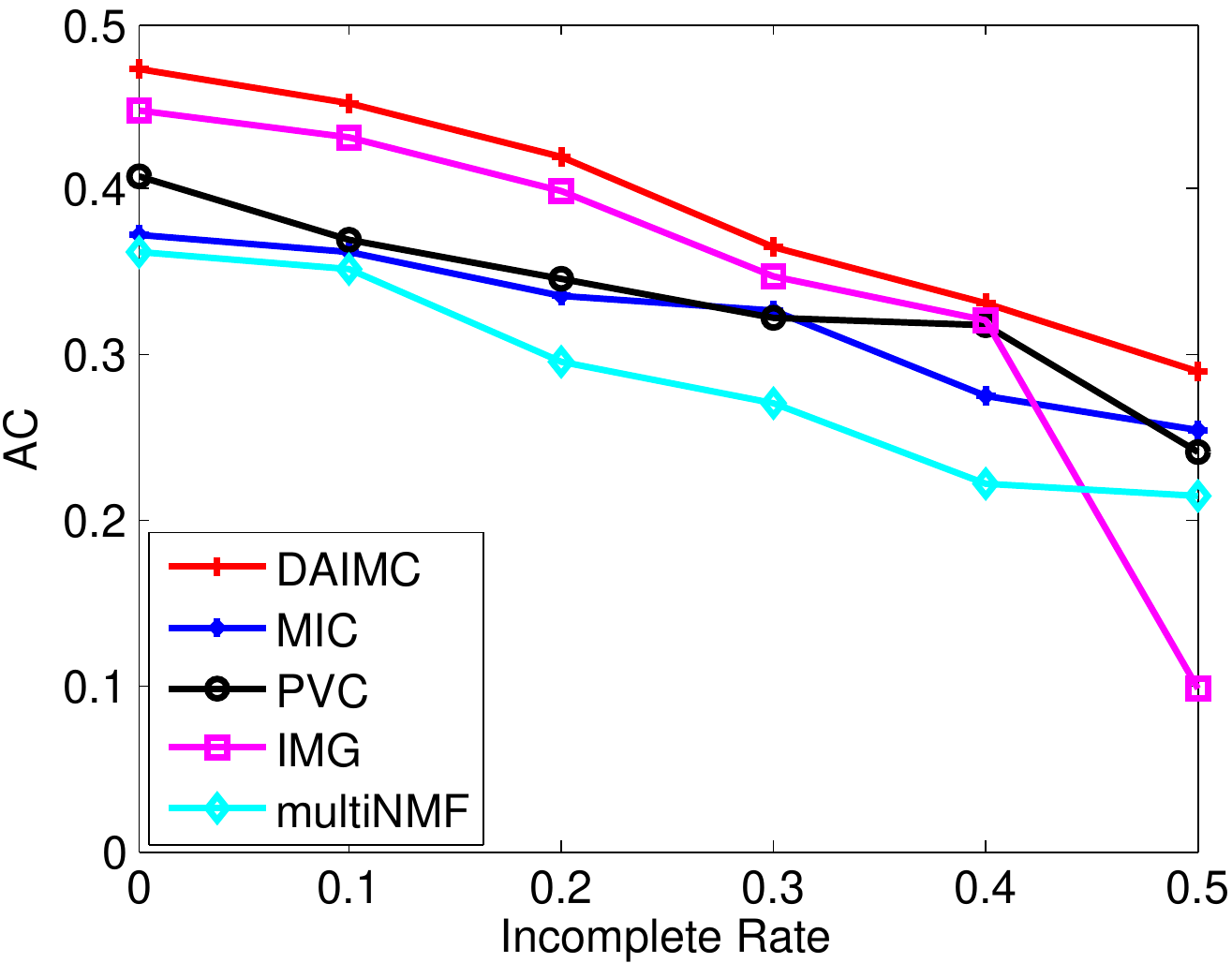}
      }
      \subfigure[NMIs for Wikipedia]{
           \label{plot:NMIs for Wikipedia}
           \includegraphics[width=0.29\textwidth]{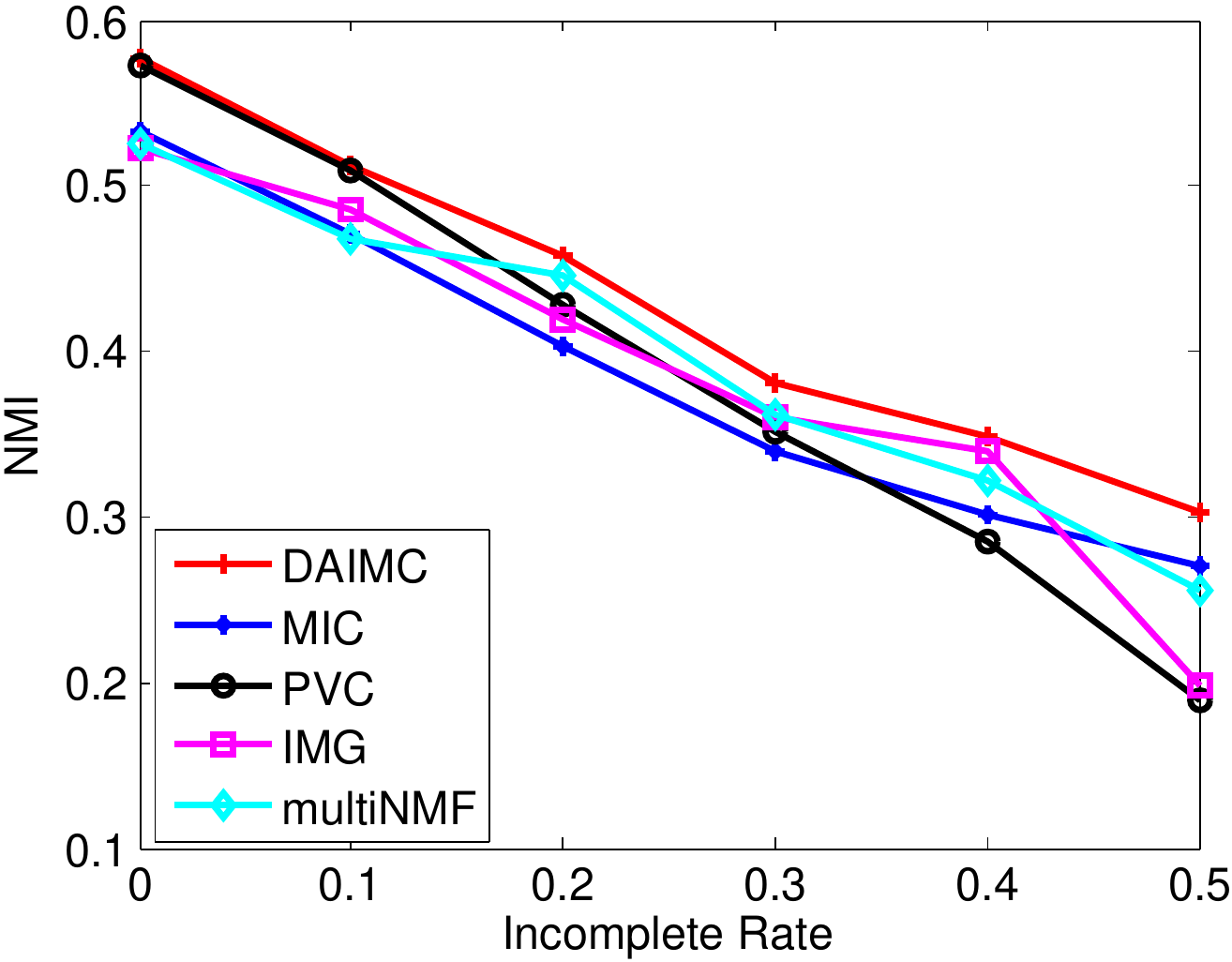}
      }
      \subfigure[NMIs for Digit]{
           \label{plot:NMIs for Digit}
           \includegraphics[width=0.29\textwidth]{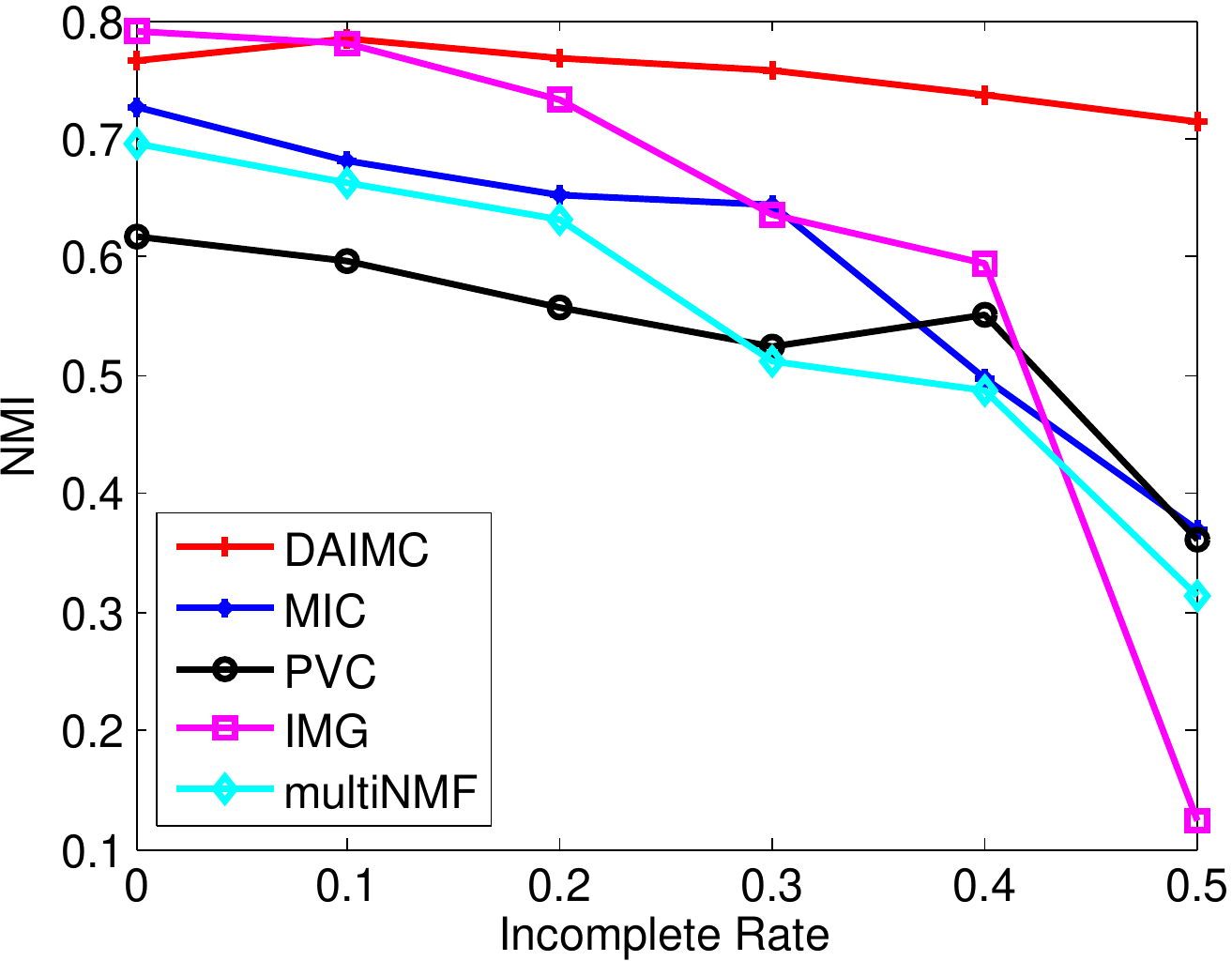}
      }
      \subfigure[NMIs for Flowers]{
           \label{plot:NMIs for Flowers}
           \includegraphics[width=0.29\textwidth]{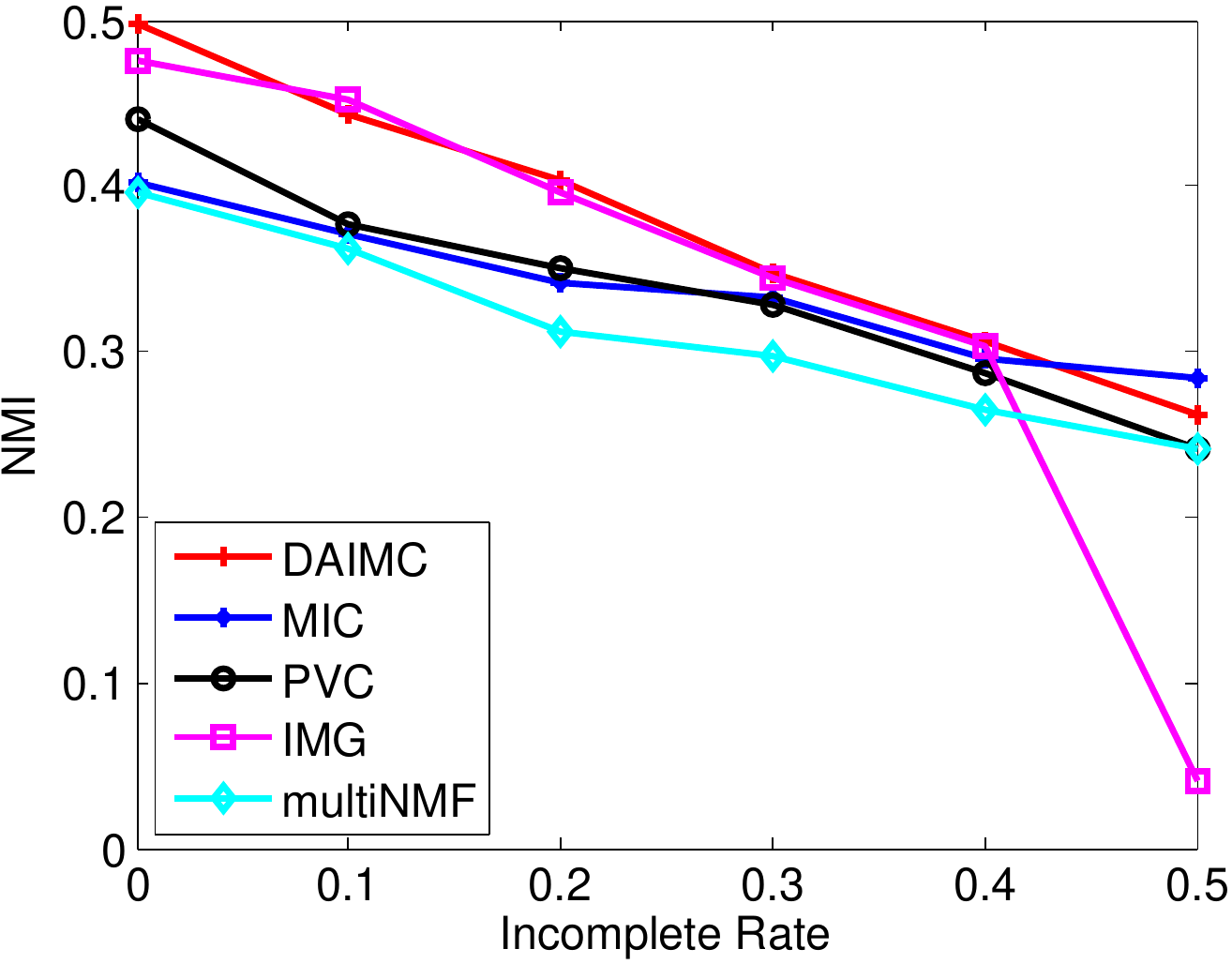}
      }
      \caption{Performance of clustering on Wikipedia, Digit and Flowers.}
      \label{Fig:NMIAC}
  \end{figure*}
$\textbf {Compared methods}$: In the experiments, DAIMC is compared with the following state-of-the-art multi-view clustering methods. (1)$\textbf {MultiNMF}$: Multi-view NMF \cite{liu2013multi} seeks a common latent subspace based on joint NMF. However, this method can not deal with the incomplete multi-view data, in our experiment, we therefore first fill the missing instances in each incomplete view with average feature values. (2)$\textbf {PVC}$: Partial multi-view clustering \cite{li2014partial} is one of the state-of-art incomplete multi-view clustering methods, which learns a common latent subspace for the aligned instances and a private latent subspace for the unaligned instances. (3)$\textbf {IMG}$: Incomplete multi-modal visual data grouping \cite{zhao2016incomplete} integrates PVC and manifold learning, which bridges the connection of missing instance data from different views by learning a complete graph Laplacian term. (4)$\textbf {MIC}$: Multiple incomplete views clustering via weighted NMF \cite{shao2015multiple} is a feasible method for incomplete multi-view clustering, which first fills the missing instances in each incomplete view with average feature values, then learns a common latent subspace with $L_{2,1}$-norm regularization. All of the hyper-parameters of these methods are selected through grid-search.

For the evaluation metric, we follow \cite{li2014partial}, using Normalized Mutual Information (NMI). Besides, precision of clustering result is also reported to give a comprehensive view. Similarly to \cite{shao2015multiple}, for the complete datasets, we randomly remove some instances from each view to make the views incomplete. The incomplete rate is from 0 (all the views are complete) to 0.5 (all the views have 50$\%$ instances missing). It is also worth to note that 3Sources is naturally incomplete. Also since PVC and IMG can only deal with two incomplete views, in order to compare PVC and IMG with other methods, we train these models on all the two-views combinations and report the best result.
\subsection{Experimental Results}
\begin{table*}
\tabcolsep 11pt
      \begin{center}
     \begin{tabular}{c|c|c|c|c|c|c|c|c}
     \hline
     \multicolumn{1}{c|}{\multirow{2}{*}{Methods}} & \multicolumn{2}{c|}{BBC-Guardian} & \multicolumn{2}{c|}{BBC-Reuters} & \multicolumn{2}{c|}{Guardian-Reuters} & \multicolumn{2}{c}{3Sources} \\ \cline{2-9}
     \multicolumn{1}{c|}{} & NMI &AC  &  NMI &AC  &  NMI &AC  & NMI& AC \\ \hline
       DAIMC&	\textbf{0.4264}&\textbf{0.5399}&\textbf{0.4482}&\textbf{0.5641}&\textbf{0.3805}&\textbf{0.5124}
       &\textbf{0.4733}&\textbf{0.5963} \\
      MIC&	0.3813& 	0.4988&	0.3814& 	0.4912&	0.3800& 	0.4612&0.4512	 &0.5631  \\
      PVC&	0.2412& 	0.4334&	0.2931&	0.4252&	0.2488& 	0.4145& $\backslash$ & $\backslash$ \\
      IMG&	0.2614&	    0.4511&	0.3612&	0.4624&	0.3411&	0.4384& $\backslash$& $\backslash$ \\
      MultiNMF&	0.3647& 0.4693&	0.3687& 0.4517&	0.3487& 0.4281&0.4134 &0.4756  \\\hline
     \end{tabular}
     \caption{The NMIs and ACs of different methods on various subsets of 3Sources.}
     \label{tab:result_3Source}
     \end{center}
\end{table*}
Table \ref{tab:result_3Source} and Figure \ref{Fig:NMIAC} report the AC and NMI values on image and text datasets with different incomplete rates, respectively. From these table and figures, we can get the following results.

From Figure \ref{plot:ACs for Wikipedia} and Figure \ref{plot:NMIs for Wikipedia}, we can see that on Wikipedia dataset, DAIMC raises the performance around  8.65$\%$ in NMI with different incomplete rate settings. And in AC, the performance of DAIMC and MIC are close for incomplete rates from 0.2 to 0.5 with the interval of 0.1, the difference between the two methods is just 1$\%$. But when the incomplete rate varies from 0 to 0.1, DAIMC outperforms all the other methods by about 3.53$\%$.

From Figure \ref{plot:ACs for Digit} and Figure \ref{plot:NMIs for Digit}, we can see that on Digit dataset, the experimental results of DAIMC are much better than those of other methods. Especially when the incomplete rates are large(0.4 and 0.5), DAIMC raises the performance around 60.78$\%$ in NMI and 64.67$\%$ in AC, respectively. The main reason for this phenomenon is due to that the Digit dataset contains 5 views. DAIMC effectively uses information from different views, reduces the impact of missing samples, and obtains better experimental results.

On Flowers dataset, from Figure \ref{plot:ACs for Flowers} and Figure \ref{plot:NMIs for Flowers}, we can easily see that DAIMC raises the performance around 10.29$\%$ in NMI and 20.37$\%$ in AC, respectively. Besides, IMG performs very well when the incomplete rate is small (0-0.4), the main reason is due to that Flowers dataset contains a terrible view $D\_texturegc$, which plays a negative role in clustering results. Thus, the performances of MIC and multiNMF are bad. In spite of this, DAIMC still performs good with the help of aligning basis matrices.

Table \ref{tab:result_3Source} shows the results on the 3Sources dataset, we conduct the experiment on all the two-views combinations and the whole dataset. From Table \ref{tab:result_3Source}, we can also observe that DAIMC outperforms all the other methods in both NMI and AC.

In summary, when dealing with text data or multi-view data  that contains less alignment information, IMG often gets poor result. Meanwhile, although MIC can handle the clustering problem with more than two-views, simply filling the missing instances with the global feature average will lead to a deviation, especially when the incomplete rate is large. By utilizing the information of instance alignment and enforcing the alignment among basis matrices, the proposed DAIMC can get better performances no matter whether it is text dataset or image dataset. Especially when the number of views is large, DAIMC yields more better results.
 \subsection{Convergence Study}
 For the convergence study, we conduct an experiment on Digit dataset with the incomplete rate of 0.4 and set the hyper-parameters $\{\alpha, \beta\}$ as $\{1e1, 1e\textnormal{-}1\}$ respectively. In Figure \ref{plot:objection}, we show the convergence curve and the NMI values with respect to the number of iterations. The blue curve shows the value of the objective function and the red dashed line indicates the NMI of our method. As can be seen, the algorithm has converged just after 30 iterations.
 \subsection{View Number Study}
 In order to demonstrate that the proposed method DAIMC can effectively exploit the information of multiple views, we conduct an experiment on Digit dataset with different view numbers. Similar to convergence study, we set incomplete rate and hyper-parameters $\{\alpha, \beta\}$ as 0.5 and $\{1e1, 1e0\}$ respectively. The results are shown in Figure \ref{plot:viewnumber}. Obviously, with the increase of the available view number, we get much better result.

  \subsection{Hyper-parameter Study}
\begin{figure}[htb]
      \centering
      \subfigure[]{
           \label{plot:objection}
           \includegraphics[width=3.9cm,height=3.0cm]{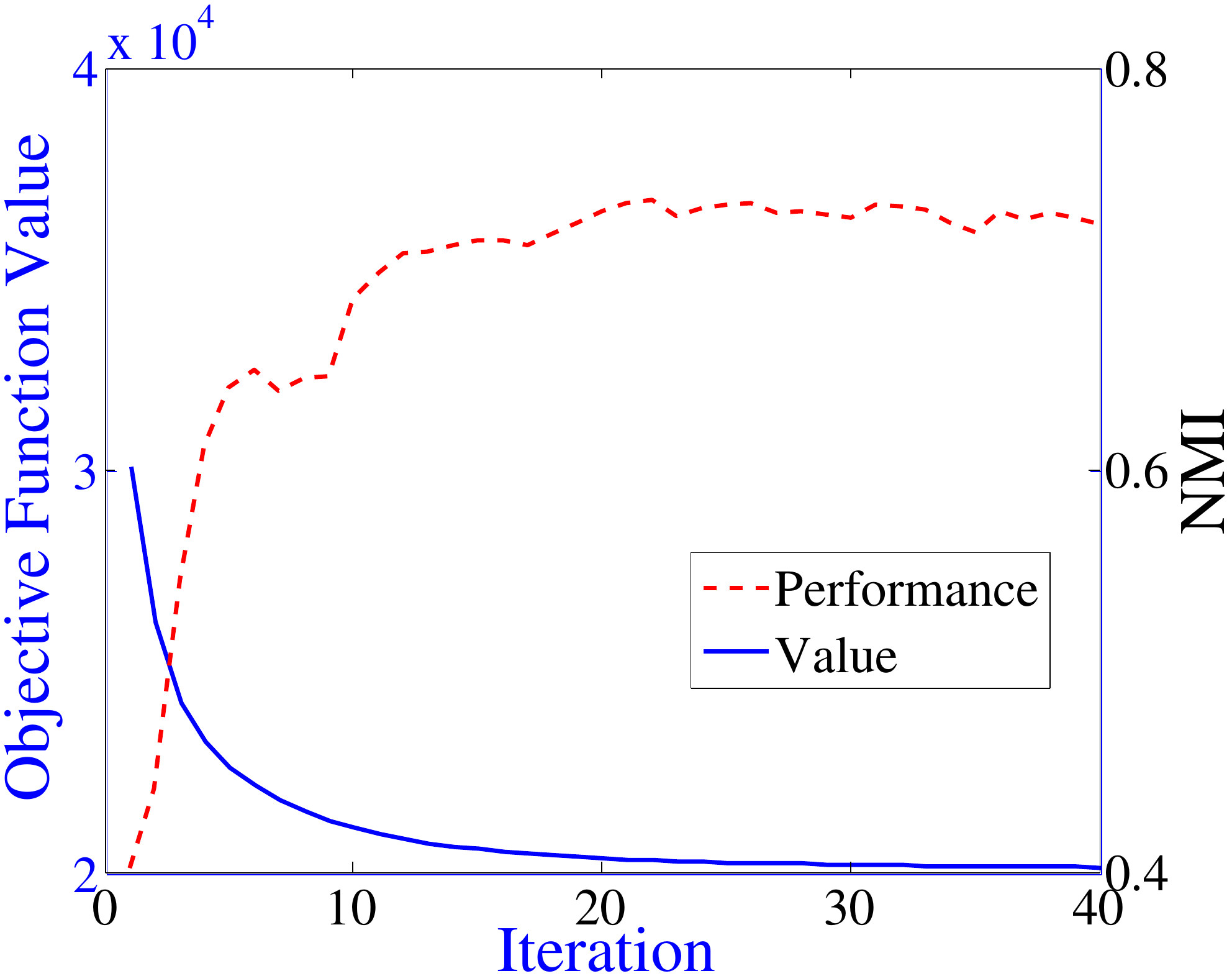}
      }
      \subfigure[]{
           \label{}
           \includegraphics[width=4cm,height=2.9cm]{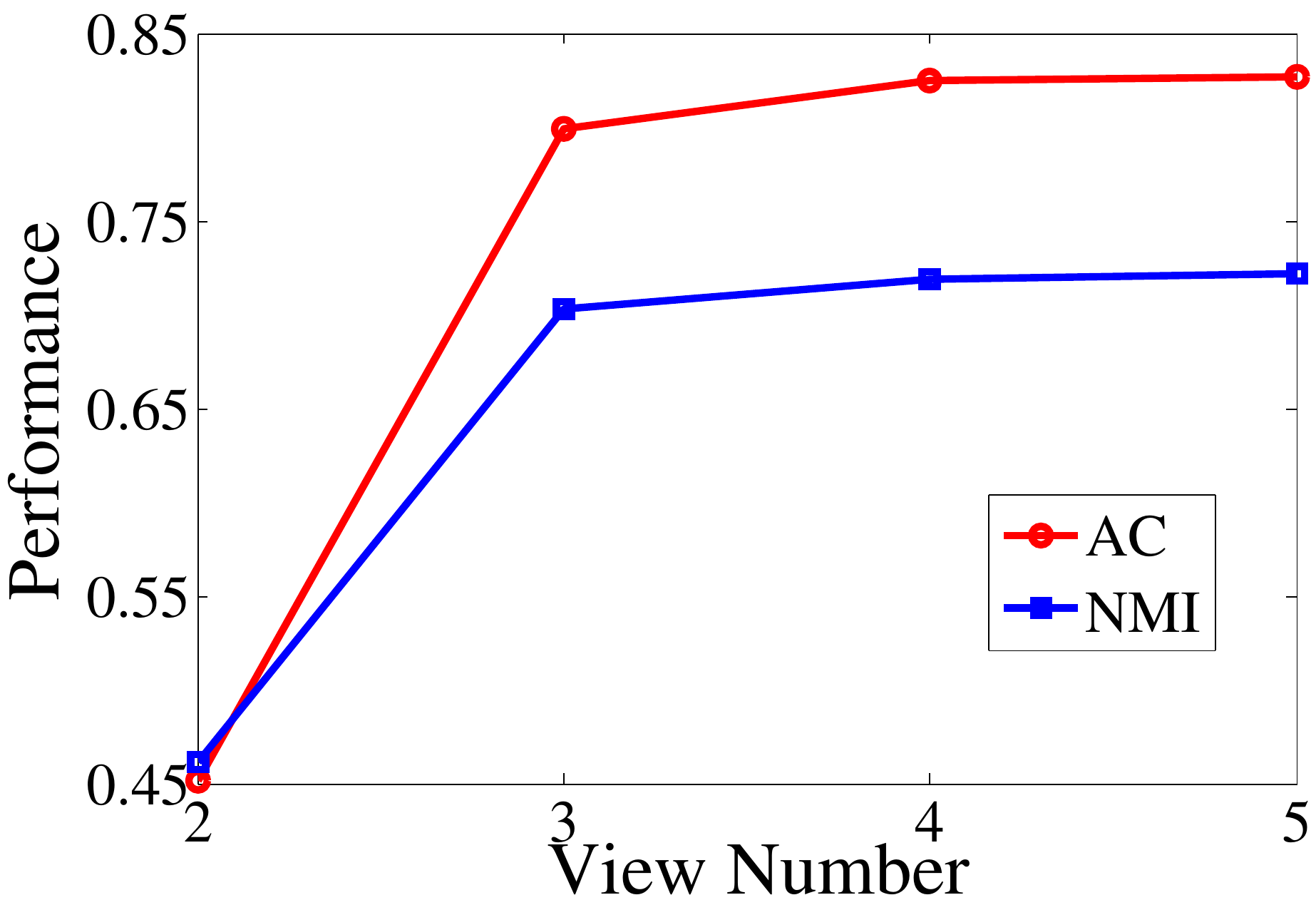}
      }
      \subfigure[]{
           \label{plot:alpha}
           \includegraphics[width=4cm,height=3.0cm]{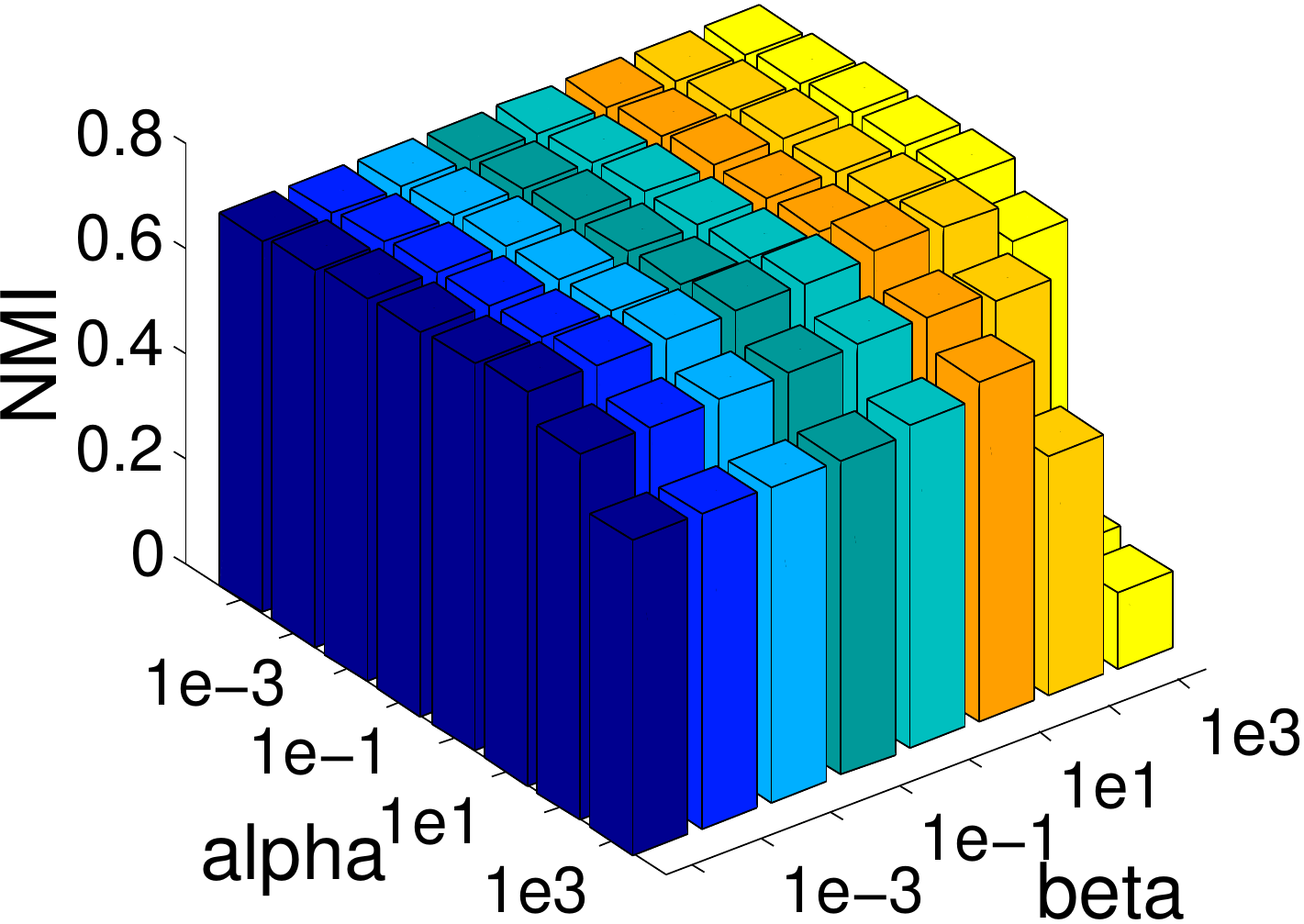}
      }
      \subfigure[]{
           \label{plot:beta}
           \includegraphics[width=4cm,height=3.0cm]{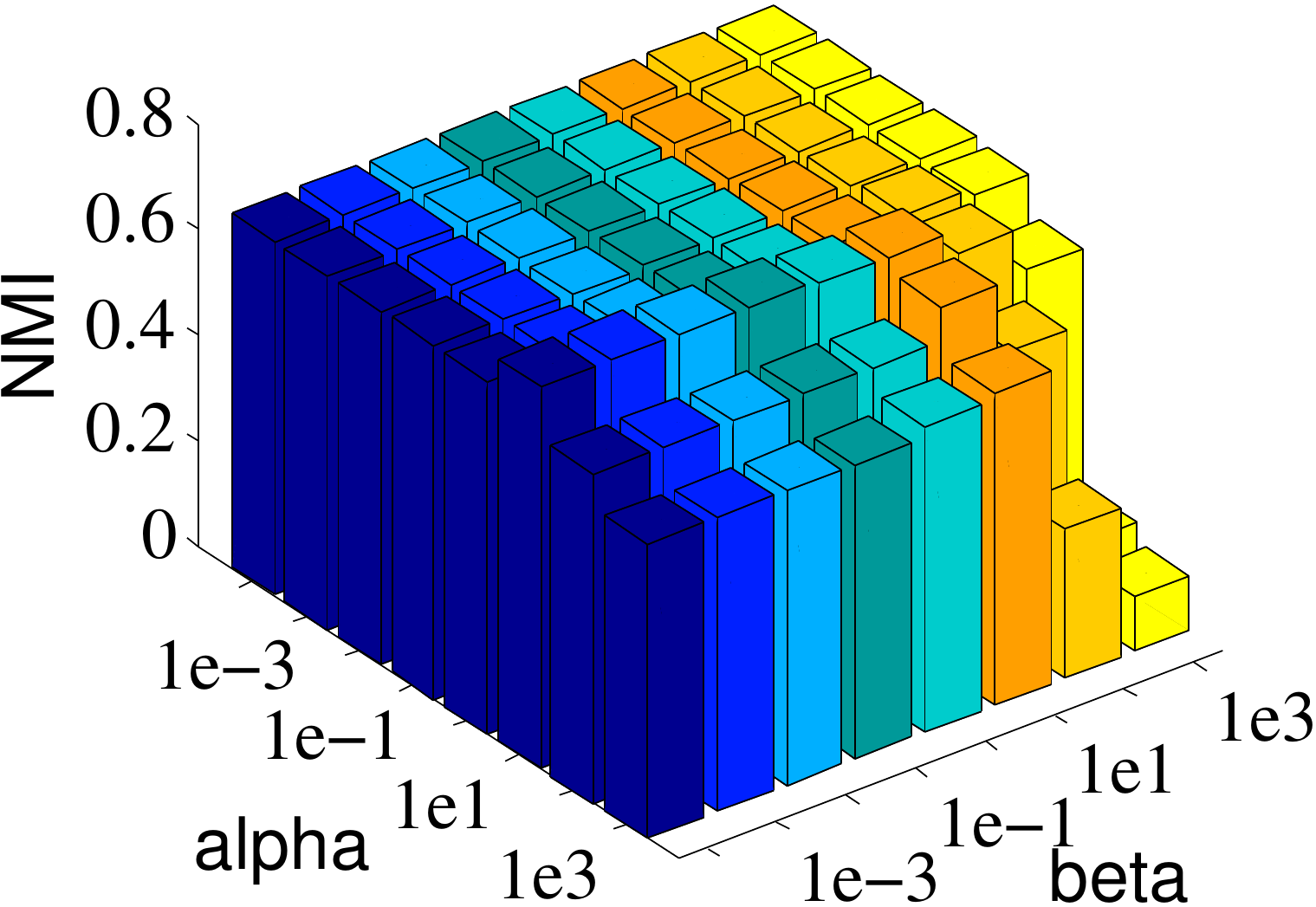}
      }
      \caption{Convergence, view number and hyper-parameter studies on the Digit dataset.}
      \label{Fig:visualization}
  \end{figure}

  The proposed DAIMC method contains two hyper-parameters $\{\alpha, \beta\}$. We conduct the hyper-parameter experiments on Digit dataset. We set the incomplete rate as 0.3 and 0.5 respectively, and report the clustering performance of DAIMC by ranging $\alpha$ and $\beta$ within the set of $\{1e\textnormal{-}4, 1e\textnormal{-}3,1e\textnormal{-}2,1e\textnormal{-}1,1e0,1e1,1e2,1e3\}$. As shown in Figure \ref{plot:alpha} and Figure \ref{plot:beta}, DAIMC obtains a relatively good performance when $\alpha=1e1$ and $\beta = \{1e\textnormal{-}1,1e0,1e1\}$.

 \section{Conclusion}
In this paper, we proposed an effective method to deal with incomplete multi-view clustering problem by considering the instance alignment information and enforcing different basis matrices being aligned simultaneously. The experimental results on four real-world multi-view datasets demonstrate the effectiveness of our method. In the future, large scale data will be considered by introducing online learning and incremental learning strategies into our model.
\newpage
\section*{Acknowledgments}
This work is supported in part by the NSFC under Grant No. 61672281, and the Key Program of NSFC under Grant No. 61732006

\bibliographystyle{named}
\bibliography{ijcai18}

\end{document}